%% file: BioKDD2025_main.tex
\documentclass[sigconf,10pt]{acmart}%natbib=false,
\usepackage{algorithm}
\usepackage{algorithmic}
\setcopyright{acmlicensed}
\copyrightyear{2025}
\acmYear{2025}
\acmDOI{XXXXXXX.XXXXXXX}
\acmConference[BIOKDD 2025]{24th International Workshop on Data Mining in Bioinformatics}{August 03,
  2025}{Toronto, Canada}

\begin{document}

\title[Counterfactual Explanations in Medical Imaging]{Counterfactual Explanations in Medical Imaging: Exploring SPN-Guided Latent Space Manipulation}

\author{Julia Siekiera}
\affiliation{%
  \institution{Johannes Gutenberg University Mainz}
  \city{Mainz}
  \country{Germany}}
\email{siekiera@uni-mainz.de}
\author{Stefan Kramer}
\affiliation{%
  \institution{Johannes Gutenberg University Mainz}
  \city{Mainz}
  \country{Germany}}
\email{kramer@informatik.uni-mainz.de}

\hypersetup{
pdfauthor={Julia Siekiera, Stefan Kramer},
}

%\renewcommand{\shortauthors}{Siekiera and Kramer}

\begin{abstract}
Artificial intelligence is increasingly leveraged across various domains to automate decision-making processes that significantly impact human lives. In medical image analysis, deep learning models have demonstrated remarkable performance. However, their inherent complexity makes them black box systems, raising concerns about reliability and interpretability. Counterfactual explanations provide comprehensible insights into decision processes by presenting hypothetical ``what-if'' scenarios that alter model classifications. By examining input alterations, counterfactual explanations provide patterns that influence the decision-making process. Despite their potential, generating plausible counterfactuals that adhere to similarity constraints providing human-interpretable explanations remains a challenge. In this paper, we investigate this challenge by a model-specific optimization approach. 
While deep generative models such as variational autoencoders (VAEs) exhibit significant generative power, probabilistic models like sum-product networks (SPNs) efficiently represent complex joint probability distributions. By modeling the likelihood of a semi-supervised VAE’s latent space with an SPN, we leverage its dual role as both a latent space descriptor and a classifier for a given discrimination task. This formulation enables the optimization of latent space counterfactuals that are both close to the original data distribution and aligned with the target class distribution. 
We conduct experimental evaluation on the cheXpert dataset. 
To evaluate the effectiveness of the integration of SPNs, our SPN-guided latent space manipulation is compared against a neural network baseline. Additionally, the trade-off between latent variable regularization and counterfactual quality is analyzed.
\end{abstract}
\begin{CCSXML}
<ccs2012>
   <concept>
       <concept_id>10010147.10010178</concept_id>
       <concept_desc>Computing methodologies~Artificial intelligence</concept_desc>
       <concept_significance>500</concept_significance>
       </concept>
   <concept>
       <concept_id>10010405.10010444.10010447</concept_id>
       <concept_desc>Applied computing~Health care information systems</concept_desc>
       <concept_significance>500</concept_significance>
       </concept>
 </ccs2012>
\end{CCSXML}

\ccsdesc[500]{Computing methodologies~Artificial intelligence}
\ccsdesc[500]{Applied computing~Health care information systems}

%\begin{CCSXML}
%TODO
%\end{CCSXML}

%\ccsdesc[500]{Do Not Use This Code~Generate the Correct Terms for Your Paper}
%\ccsdesc[300]{Do Not Use This Code~Generate the Correct Terms for Your Paper}
%\ccsdesc{Do Not Use This Code~Generate the Correct Terms for Your Paper}
%\ccsdesc[100]{Do Not Use This Code~Generate the Correct Terms for Your Paper}

\keywords{Sum-Product Networks, Explainable AI, Medical Imaging}

%\received{15 May 2025}


\maketitle
%\onecolumn

\section{Introduction}

\begin{figure*}%[h!]

\begin{center}

\includegraphics[width=\textwidth]{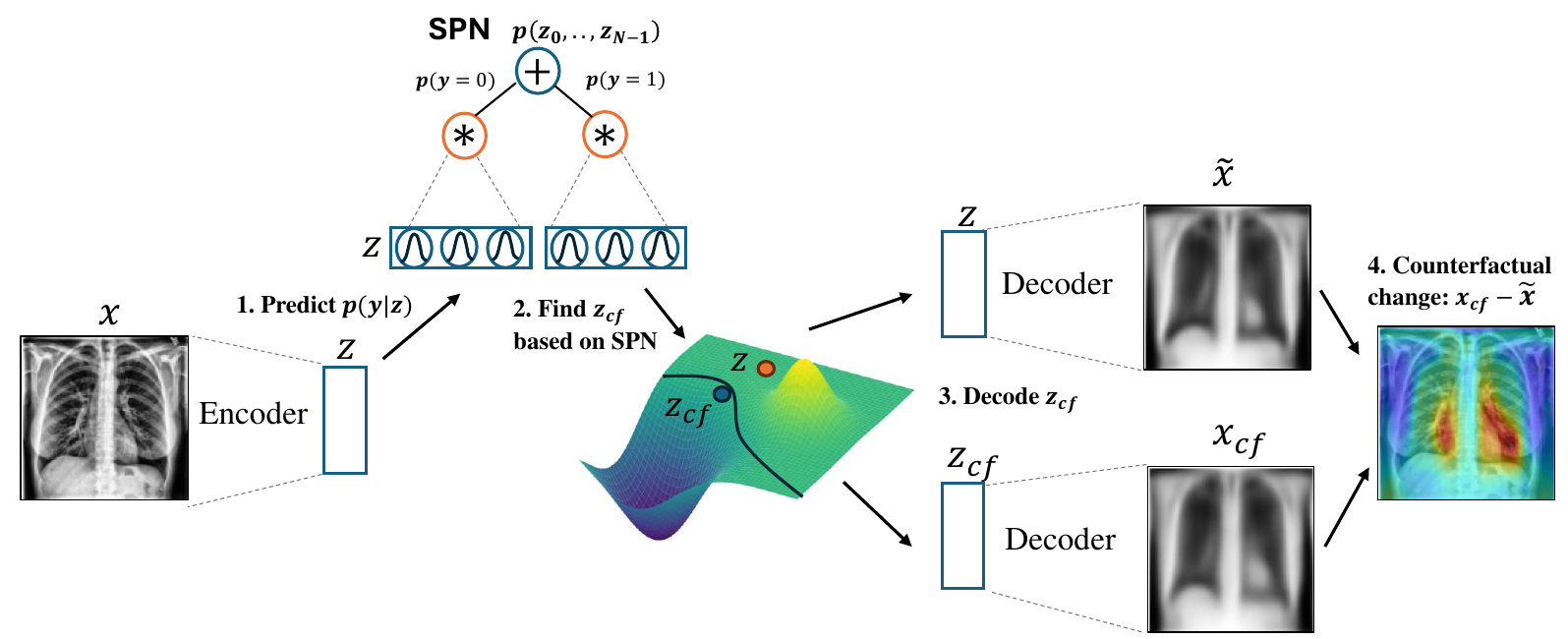}
\caption{Overview of the individual steps in our proposed method to generate counterfactual explanations. A gaussian SPN is created to describe the embedding space $z$ of a trained VAE to find plausible counterfactual embeddings within this space. The counterfactual changes can be visualized with the decoder as predictive explanations.}
\label{fig:overview}
\end{center}
\end{figure*}
Deep neural networks (DNNs) have demonstrated remarkable success in various domains due to their ability to identify complex patterns in high-dimensional data. However, their inherent complexity turns them into black-box systems, raising significant concerns regarding reliability and interpretability. This is particularly critical in sensitive applications such as medicine, where understanding model decisions is essential for clinical validation and trustworthiness. This lack of transparency hinders their deployment in real-world scenarios such as in medical imaging for disease diagnosis. Enhancing interpretability in this domain can help to better understand model predictions, detect potential biases, and ultimately improve patient care. A promising research direction within the field of explainable artificial intelligence (XAI) are counterfactual explanations, which employ hypothetical ``what-if'' scenarios to illustrate the decision-making process of a model. The objective is to modify an existing input of a classifier in such a way that the model assigns it a different classification. This approach enables the identification of relevant features that the model recognized for decision-making. Counterfactual explanations are particularly beneficial in medical imaging, as they allow practitioners to analyze which image features influence a diagnosis and how slight variations might alter the predictions.

Sum-product networks (SPNs)~\cite{6130310} are probabilistic models designed to efficiently represent joint probability distributions by constraining their internal structure. They exhibit several notable advantages, making them a promising approach in probabilistic modeling. SPNs allow for exact and efficient inference, even in high-dimensional and complex probabilistic models. They are particularly known for their representation capabilities, enabling them to model complex probability distributions compactly. Additionally, SPNs inherently support uncertainty estimation, making them valuable in tasks where probabilistic reasoning is crucial. SPN learning allows not only to optimize the internal weights, but even the entire graph structure can be learned, enabling a deeper understanding of the underlying probability distribution by revealing dependencies and hierarchical structures within the data. Despite these advantages, SPNs have received limited attention in the deep learning community and remain less explored compared to other machine learning paradigms. However, recent efforts to map and train SPNs as constrained neural networks have shown promising results~\cite{pmlr-v115-peharz20a,Butz_Oliveira_dos_Santos_Teixeira_2019,pmlr-v138-wolfshaar20a}. Extending this line of research, we explore a hybrid approach that combines DNNs with SPNs, leveraging the strengths of pattern recognition with the ability to represent exact probability distributions in the context of counterfactual generation.

We propose an SPN-guided approach within a (semi-) supervised Variational Autoencoder (VAE), which learns latent representations with respect to both the generative process and a given classification task for counterfactual generation. Counterfactual examples are created by manipulating the latent space of the VAE. We explore how SPNs can enhance this process by %replacing the neural network classifier with an SPN and U
describing the latent space with respect to the classification task to introduce additional constraints on counterfactual generation that ensure plausible and meaningful alterations. An overview of our proposed method can be found in Figure~\ref{fig:overview}.

Our contributions are the following:
\begin{enumerate}
    \item We propose the integration of a structure-learning SPN into the latent representation of a VAE.
    \item We demonstrate that the SPN can act as classifier and descriptor of the latent space, enabling the generation of counterfactual examples on medical image data. %by controlling the plausibility of the latent vector.
    \item We investigate the relationship between the complexity of the latent variable representation and the quality of counterfactual examples, providing insights into the trade-offs involved.
\end{enumerate}
The source code of our method and experimental setup is available at \href{https://github.com/kramerlab/SPN_LSM}{https://github.com/kramerlab/SPN\_LSM}.

\section{Related Work}

Recent research has explored the integration of Sum-Product Networks (SPNs) with deep learning to leverage the strengths of both methodologies. Several studies have implemented SPNs within deep learning frameworks such as TensorFlow, enabling more flexible and efficient training. For example, RAT-SPN~\cite{pmlr-v115-peharz20a} introduces a randomized prior structure, while Convolutional SPNs~\cite{Butz_Oliveira_dos_Santos_Teixeira_2019,pmlr-v138-wolfshaar20a} incorporate convolutional operations constrained by SPN principles. These hybrid architectures benefit from the tractable probabilistic inference capabilities of SPNs and the representational power of DNNs, making them particularly suitable for image processing. Additionally, the framework SPFlow~\cite{molina2019spflow} is able to convert SPNs generated by structure learners into deep learning-compatible formats to fine-tune the estimated weights inside the SPNs structure. Other studies employ SPNs as probabilistic descriptors within deep learning models, such as for active uncertainty learning~\cite{khosravani2024using}, where SPNs are applied to feature representations to derive exact probabilities for uncertainty sampling. 

Despite these advances, relatively few studies have investigated the role of SPNs in XAI. Veragi et al.~\cite{veragi_understanding_spns} focused on improving interpretability of SPN structure learners for tabular data, such as LearnSPN~\cite{pmlr-v28-gens13}, by generating understandable representations based on structure and node values. However, the scalability of these representations is limited by the number of nodes, making them impractical for large networks. Converting SPNs to context-specific independence trees (CSI-trees)~\cite{pmlr-v186-karanam22a} that are both interpretable and explainable to domain experts, as well as the solution to generate likely counterfactuals~\cite{nvemevcekgenerating} by integrating SPNs into mixed-integer optimization, are both limited to lower dimensional input data. 

Counterfactual explanations have gained significant traction in deep learning, particularly for image data, as they provide intuitive modifications that alter model predictions. An overview of counterfactual explanations and their benchmarking is provided in a paper by Guidotti~\cite{guidotti2024counterfactual}, which discusses various theoretical and algorithmic perspectives. A key challenge in counterfactual image generation is ensuring that the generated images remain plausible, while preserving the similarity to the original instance but changing the model decisions at the same time.

Generative adversarial networks (GANs) have been widely applied for realistic counterfactual image generation, such as in StyleGAN-based approaches~\cite{atad2022chexplainingstylecounterfactualexplanations} and conditional GAN (cGAN) frameworks~\cite{SINGLA2023102721}, which perturb images to flip classification decisions while preserving anatomical structure and foreign objects. However, GAN-based methods often struggle to meet the similarity criterion required for meaningful counterfactual explanations. Jeanneret et al.~\cite{jeanneret2023adversarial} propose transforming adversarial attacks into semantically meaningful perturbations without modifying the classifier. Their approach leverages denoising diffusion probabilistic models (DDPMs) as regularizer to prevent high-frequency and out-of-distribution perturbations in adversarial attacks. Additionally, latent diffusion models have been explored for concept discovery in counterfactual reasoning~\cite{varshney2025generating}, though these methods rely on manual latent space exploration.

Several approaches investigated autoencoder architectures to manipulate latent space representations, which offer more controllability in ensuring minimal and plausible counterfactual modifications. For instance, deterministic autoencoders structure latent spaces by class labels to generate representative counterfactuals~\cite{10.1007/978-3-030-93736-2_10}, while gradient-based latent vector shifts~\cite{pmlr-v143-cohen21a} provide an alternative approach. Gaussian discriminant VAEs (GDVAEs)~\cite{10.1007/978-3-031-73668-1_18} have also been proposed for end-to-end counterfactual generation. Moreover, hierarchical VAEs (HVAEs) combined with deep structural causal models~\cite{pmlr-v202-de-sousa-ribeiro23a} have been proposed to estimate high-fidelity counterfactuals. In our work, we aim to further investigate the promising approach of counterfactual generation in the latent space of VAEs. Specifically, we explore the integration of SPNs to describe the latent space to facilitate the automated discovery of counterfactuals. 
We examine the trade-off of different objectives to optimize this process.

\section{Methods}
\subsection{Variational Autoencoders}

Variational Autoencoders (VAEs)~\cite{kingma2013auto, rezende2014stochastic} are generative models that learn a latent data representation by leveraging probabilistic inference. They offer a flexible and principled framework for learning structured representations, making them a powerful tool for the generation of counterfactual examples. Unlike traditional autoencoders, which rely on deterministic encoding-decoding mechanisms, VAEs introduce a probabilistic framework that enables the generation of new data samples from the learned latent distribution.
VAEs consists of an encoder network, a decoder network, and the latent space. Given a sample $x$, the encoder parameterizes an approximate posterior distribution $q_\phi(z|x)$ over the latent variables $z$. The decoder $p_\theta(x|z)$ reconstructs the input by sampling from the latent space. 
VAEs maximize the evidence lower bound (ELBO) on the marginal log-likelihood of the data:
\begin{equation*}
%\label{eq:elbo}
    \log p_\theta(x) \geq \mathbb{E}_{q_\phi(z|x)} [\log p_\theta(x|z)] - D_{KL}(q_\phi(z|x) || p(z)).
\end{equation*}
The first term controls the reconstruction error, while the second term, the Kullback-Leibler divergence (KLD) between the approximate posterior $q_\phi(z|x)$ and the prior $p(z)$, regularizes the latent space $q_\phi(z|x)$.
In its original form, the VAE was proposed for unsupervised learning, but they can be extended to semi-supervised learning by incorporating label information into the latent space~\cite{kingma2014semi}. 
We build upon this idea in our proposed VAE architecture to create a latent space $q_\phi(z|x)$ that contains not only information of the generative process, but whose features are also suitable for a given classification task. Therefore, we add a classifier $q_{\phi_1}(y|z)$ that aims to predict the class labels based on the latent representation. 
The ELBO is extended by the classification performance to obtain a three termed loss function that we aim to minimize:
\begin{align}
\label{eq:elbo_extended}
\mathcal{L} =\ & \beta_{0} \cdot -\mathbb{E}_{q_\phi(z|x)} [\log p_\theta(x|z)] \nonumber \\
& + \beta_{1} \cdot D_{KL}(q_\phi(z|x) \,\|\, p(z)) + \beta_{2} \cdot \mathcal{H}(q_{\phi_1}(y|z))
\end{align}

with $\mathcal{H}$ representing the cross-entropy.
Inspired by the beta-VAE~\cite{higgins2017beta}, the trade-off between the three objectives, data reconstruction, classification and KLD regularization, is controlled by constant weights $\beta_0,\beta_1\text{ and }\beta_2\in \mathbb{R}$ to investigate their corresponding trade-off.

\subsection{Sum-Product Networks (SPN)}
% 1. SPN in general
SPNs, introduced by Poon and Domingos~\cite{6130310}, are an extension of Darwiche's arithmetic circuits~\cite{Darwiche2002ALA,10.1145/765568.765570}. The primary innovation lies in their structural constraints, specifically completeness and decomposability, which facilitate exact and efficient probabilistic inference while preserving tractability. SPNs are probabilistic models designed to represent joint probability distributions efficiently. Given a set of random variables $ \{ X_i \}_{i=1}^N $ defined over a domain $ \mathbf{X} $, an SPN aims to estimate the joint probability distribution $ p_\mathbf{X} $. Structurally, an SPN is a rooted acyclic directed graph where each node or sub-SPN $S_i$ corresponds to a probability distribution. The leaf nodes $ n $ represent univariate probability distributions, whereas internal nodes perform either product or weighted sum operations over their child distributions.

By enforcing structural constraints such as completeness and decomposability, SPNs enable efficient and tractable probabilistic inference. These constraints differentiate SPNs from traditional graphical models and contribute to their scalability for large-scale datasets. Formally, let $ S $ denote an SPN, with $ S_\oplus $ and $ S_\otimes $ representing the sets of sum and product nodes, respectively. 

The scope of a node $ n $ in $ S $ is defined as:
\begin{equation*}
    \mathbf{X} \supseteq \text{sc}(n) = \bigcup_{c \in \text{ch}(n)} \text{sc}(c)
\end{equation*}
with $ \text{ch}(n) $ the children of a node $ n $.
an SPN $ S $ is complete iff:
\begin{equation*}
    \forall n \in S_\oplus, \, \forall c_1, c_2 \in \text{ch}(n): \text{sc}(c_1) = \text{sc}(c_2).
\end{equation*}
an SPN $ S $ is decomposable iff:
\begin{equation*}
    \forall n \in S_\otimes, \, \forall c_1, c_2 \in \text{ch}(n), \, c_1 \neq c_2: \text{sc}(c_1) \cap \text{sc}(c_2) = \emptyset.
\end{equation*}
If $ S $ satisfies both completeness and decomposability, it is valid~\cite{6130310}. A valid SPN allows for efficient evaluation of the joint unnormalized probability distribution $ p_\mathbf{X} $~\cite{pmlr-v38-peharz15}. One of the most prominent methods to learn this structure top down from the data is LearnSPN~\cite{pmlr-v28-gens13}.

Despite their advantages, SPN structure learners face challenges regarding training efficiency, particularly in the high-dimensional data domain. Integrating SPNs into neural network architectures offers a novel approach to leveraging the strengths of both methods. We propose an architecture that employs a convolutional VAE to transform the high-dimensional input into a lower-dimensional, controllable latent space. The latent representation $(z,y)$ is subsequently processed  by LearnSPN~\cite{pmlr-v28-gens13}, which recursively partitions the data to construct a structure that closely aligns with the underlying data distribution. Since this structure is learned directly from the data, it captures statistical dependencies that generalize well to unseen data, given enough training examples. To enable counterfactual optimization via gradient backpropagation, we transfer the learned SPN structure into a neural network-based framework. We worked with SPFlow~\cite{molina2019spflow} to execute LearnSPN and updated the existing SPN-to-TensorFlow 1 converter for compatibility with TensorFlow 2. Additionally, we extended the library to support discriminative models in TensorFlow.

\subsection{SPN-guided counterfactual explanations}
\begin{figure}[t!]%[htb]

\begin{center}
\includegraphics[width=0.48\textwidth]{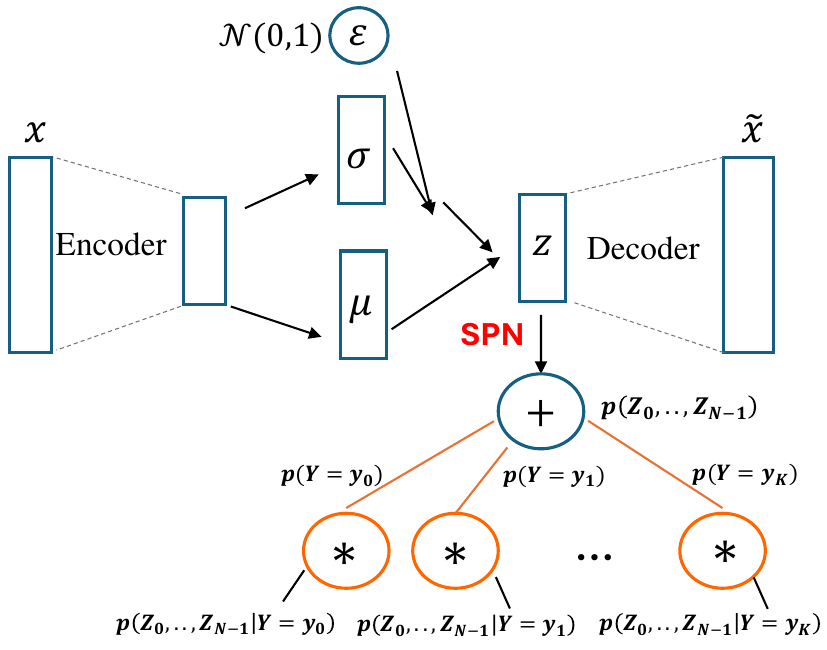}
\caption{Model architecture for SPN-guided counterfactual explanations with VAEs. The VAE learns a regularized latent representation $z$ that is suitable for both classification and input reconstruction. The likelihood of $z$ is described by an SPN whose sub-SPNs of the root node represent the distribution of $z$ given a certain class label $y$.}
\label{fig:VAE_SPN}
\end{center}

\end{figure}

Our model shown in Figure~\ref{fig:VAE_SPN} integrates an SPN into the architecture of a (semi-)supervised VAE to enable targeted latent space manipulation for the generation of counterfactual examples, satisfying minimality and plausibility. Instead of searching for counterfactual examples in the input space $X$, we generate them by manipulating the latent space of a VAE. By this approach, we benefit from the robustness of latent space representations and their probabilistic interpretability that can be efficiently described by SPNs.

Our proposed method involves the execution of three steps. In the first step, we train a VAE with an integrated classifier to learn a latent feature space, which represents the most meaningful characteristics for the generation of new instances and for a given classification task. The en- and decoder of the VAE are modeled by convolutional (ResNet CNN) layers, while the classifier $p(c|z)$ is initially modeled by a multilayer perception (MLP). The initial classification MLP ensures fast training of an encoder that is able to capture not only generation-relevant features but also classification-relevant features. 

After VAE training, the MLP classifier $p(c|z)$ is replaced by an SPN. The SPN structure is learned and integrated into the neural network by our extended TensorFlow converter, %of a TensorFlow converter~\cite{molina2019spflow}, 
assuming that each feature $z_i$ follows a Gaussian distribution. The classification SPN is modeled by the distribution of its sub-SPNs describing the latent space distribution for each class:
\begin{align*}
p_{spn}(Y = y_0 \mid Z_0, \ldots, Z_{N-1}) 
&= 
\nonumber \\
&\hspace{-6em} \frac{
p_{spn}(Z_0, \ldots, Z_{N-1} \mid Y = y_0) \cdot p_{spn}(Y = y_0)
}{
\sum\limits_k p_{spn}(Z_0, \ldots, Z_{N-1} \mid Y = y_k) \cdot p_{spn}(Y = y_k)
}
\end{align*}
The distribution of the latent variables $p_{SPN}(z)$ is described by the weighted sum of the sub-SPNs.

In the last step, we generate for each instance $x$ with class label $y$ the counterfactual $x_{cf}$ that will switch the classifier's prediction to the class $y_{cf}$. A summary of the individual steps is given in Algorithm~\ref{algo:cf}. We sample $z \sim p(z|x)$ $r$ times from the encoder. The sampled latent representation represents the baseline for the counterfactual manipulation $z_{cf}$. Unlike other approaches that manipulate each variable of the latent representation independently from each other~\cite{varshney2025generating}, we manipulate all variables simultaneously by gradient descent optimization. This approach benefits from less requirements regarding the latent representation as classification decisions could also depend on multivariate features.
Our investigated latent space objective:
\begin{align}
\label{eq:clf}
    z_{cf} =\ & \underset{z'}{\text{arg max}} \; \log(p(y_{cf} | z')) - \beta ||z' - z||^2 
    \nonumber \\
& - \gamma |\log(p(z')) - \log(p(z))|
\end{align}
includes three different target constrains. An optimal $z_{cf}$ should increase the class probability of $y_{cf}$ with minimal changes regarding the original $z$ ensuring that the likelihood of $z$ and $z_{cf}$ remains similar. The two latter terms that aim to control the minimality and the plausibility aspect of the counterfactuals are weighted by $\beta,\gamma\in\{0,1\}$ to evaluate the impact of the SPN-guided alterations.
The resulting latent representations $z_{cf}$ are decoded to generate the corresponding $x_{cf}$. 

\begin{algorithm}

\caption{Generation of SPN-guided counterfactual examples}
\begin{algorithmic}[1]
    \STATE \textbf{Input:} image $x$; number of replicates $R$; VAE: $z = \text{Enc}(x)$, $\tilde{x} = \text{Dec}(z)$; \text{SPN:} $p_{spn}$
    \STATE $X \leftarrow (x,...,x)\in \mathbb{R}^{R\times h\times w}$
    \STATE Sample: $z \leftarrow \text{Enc}(X)$
    \STATE $z_{cf} \leftarrow z$
    \STATE Optimize $z_{cf}$ with SGD:
    \begin{itemize}
        \item $z_{cf} = \underset{z'}{\text{arg max}} \; \log(p_{spn}(y_{cf} | z')) - \beta ||z' - z||^2 - \gamma |\log(p_{spn}(z')) - \log(p_{spn}(z))|$
        %\item \textbf{MLP:} $z_{cf} = \underset{z'}{\text{arg max}} \; p(y_{cf} | z') - \beta ||z' - z||^2$
    \end{itemize}

    \STATE $x_{cf} \leftarrow  \frac{1}{r}\sum_{r\in R} \text{Dec}(z_{cf_{r}})$
    \STATE $\tilde{x}\leftarrow  \frac{1}{r}\sum_{r\in R}  \text{Dec}(z_r)$
    
\end{algorithmic}
\label{algo:cf}
\end{algorithm}

\section{Evaluation}
\subsection{Experimental Setting}

The evaluation of our proposed method is conducted on the CheXpert dataset~\cite{10.1609/aaai.v33i01.3301590}\footnote{The CheXpert dataset was retrieved from a downsampled version available at \url{https://huggingface.co/datasets/danjacobellis/chexpert}}. To mitigate confounding factors such as multiclass classification complexity and class imbalance, we focus on a binary classification scenario with balanced class distribution. Specifically, we filter the dataset to include only instances labeled as ‘cardiomegaly’ and ‘no finding’ and adjust the class distribution to a 50/50 ratio by downsampling the majority class.

Preprocessing steps include contrast enhancement with Contrast Limited Adaptive Histogram Equalization (CLAHE) and image rescaling to a resolution of 128×128 pixel. The dataset is partitioned into 80\% training and 20\% test set. Within the training set a three-fold cross-validation procedure is applied resulting into 21,572 training, 5,532 validation and 6,896 test instances. We ensured that images with the same patient identifier are not distributed across the 3 datasets.
Our presented results are the mean over all splits.

%neural network configurations
For the VAE, we chose the following hyperparameters: ReLU as activation function, the KLD regularizer $\beta_{1}\in \{0.1,\allowbreak 0.01,\allowbreak 0.001,\allowbreak 0.0001\}$, a batchsize of 50, a latent $z$ dimension of 62, and ADAM as optimizer with a learning rate of 0.001 for 100 epochs. 
The en- and decoder architecture were implemented with a CNN ResNet architecture of  7 layers (with 16, 32, 62, 124, 124, 124, 124 as number of filters) and Gaussian noise of $\mu=0$ and $\sigma=0.2$ for each layer.

To investigate the impact of the objective in Equation~\ref{eq:clf} for counterfactual generation, we generate counterfactuals for $\beta\in\{0,1\}$, $\gamma\in\{0,1\}$ with an SPN and an MLP classifier. The MLP classifier with $\gamma\in\{0,1\}$ serves as a baseline for standard neural network performance.

\subsection{Evaluation metrics}
In general, the evaluation of counterfactual explanations is methodologically challenging due to the absence of ground truth data and the different requirements imposed on counterfactual explanations.
In this work, we focus on established metrics that capture distinct and well-documented aspects of counterfactual quality~\cite{guidotti2024counterfactual}: validity, proximity, and plausibility. 
Validity, the change of the model prediction, is commonly measured with the flip rate, i.e. how many counterfactuals are classified as the targeted label.
\begin{equation}
    \text{Validity} = \frac{1}{|X|}\sum_{x\in X}1_{[f(x)\neq f(x_{cf})]}
\end{equation}
Proximity quantifies the distance between a counterfactual and the original input instance. To measure this, the L2 norm between $x$ and $x_{cf}$ is considered. 
The plausibility of counterfactual explanations can be assessed with the Fréchet Inception Distance (FID)~\cite{heusel2017gans}:
\begin{equation}
    \text{FID} = \|\mu_{org} - \mu_{cf}\|^2 + \text{Tr}[\Sigma_{org} + \Sigma_{cf}] - 2\text{Tr}[\Sigma_{org}\Sigma_{cf}]
\end{equation}
Generally, FID evaluates generative models by comparing the statistics of embeddings from the Inception V3 network~\cite{szegedy2016rethinking} from the original and the generated dataset. In the context of counterfactuals, this score quantifies the degree to which counterfactual instances resemble samples from the original dataset. FID is defined based on the mean Inception embeddings, $\mu_{org}$ and $\mu_{cf}$, and covariance matrices, $\Sigma_{org}$ and $\Sigma_{cf}$, corresponding to the original dataset and the generated counterfactuals, respectively.
Lower FID corresponds to better image quality. To match the inputs of the Inception V3 model (299 × 299 pixels), we rescale the original $x$ as well as the counterfactuals $x_{cf}$.
Furthermore, the number of training epochs needed to change the models classification prediction for successful counterfactuals is reported.

\subsection{VAE Performance}
To interpret the results of counterfactual generation, we first discuss the three combined objectives of Equation~\ref{eq:elbo_extended}: reconstruction quality, latent space regularization, and classification performance. Table~\ref{table:VAE_performance} %and~\ref{table:1} 
presents the reconstruction performance, measured by the mean absolute error (MAE) and mean squared error (MSE), under different values of the KLD regularization $\beta_{1}$. As expected, for both datasets, lower $\beta_{1}$ leads to less KLD regularization (KLD increases), while MAE and MSE decrease. These findings highlight the inherent trade-off between regularization strength and reconstruction quality, modulated by $\beta_{1}$.
\begin{table}%[h!]
\caption{VAE performance for different $\beta_{1}$. We show the influence of trading-off latent space generalization with KLD and reconstruction performance measured by MAE and MSE. Best results are displayed as bold. }% (test set 1)}
\label{table:VAE_performance}
\centering
\begin{tabular}{c|ccc}
\toprule
$\beta_{1}$ &   MAE &   MSE &     KLD \\
\midrule
0.1000 & 0.157 & 0.039 &   \textbf{1.932} \\
0.0100 & 0.129 & 0.028 &  15.894 \\
0.0010 & 0.104 & 0.019 &  86.738 \\
0.0001 & \textbf{0.098} & \textbf{0.017} & 252.265 \\

\bottomrule
\end{tabular}

\end{table}

The classification performance of both the MLP baseline and the SPN is presented in Table~\ref{table:VAE_classification}. 
By varying $\beta_{1}$ the trade-off between KLD regularization and classification performance gets visible. 
Only for $\beta_{1}=0.1$, the SPN achieves an improvement in classification performance. However, for smaller $\beta_{1}=0.1$, the MLP outperforms the SPN. Notably, in terms of entropy, the MLP baseline yields significantly lower values, suggesting that the MLP produces more confident predictions with lower uncertainty compared to the SPN.
\begin{table*}%[h!]
\caption{Classification performance of the MLP baseline classifier and the SPN for different $\beta_{1}$. Best results are displayed as bold.}
\label{table:VAE_classification}
\centering
\begin{tabular}{cc|ccccc}
\toprule

Classifier & $\beta_{1}$ &  Accuracy &  Entropy &   AUC &  Precision &  Recall \\
\midrule
MLP &  0.1000 &     0.826 &    1.314 & 0.894 &      0.834 &   0.812 \\
SPN &  0.1000 &     0.830 &    1.035 & 0.897 &      0.827 &   0.832 \\
MLP &  0.0100 &     0.845 &    0.365 & \textbf{0.919} &      \textbf{0.870} &   0.810 \\
SPN &  0.0100 &     0.834 &    0.756 & 0.915 &      0.836 &   0.829 \\
MLP &  0.0010 &     0.849 &    \textbf{0.359} & 0.917 &      0.867 &   0.823 \\
SPN &  0.0010 &     0.831 &    1.536 & 0.902 &      0.809 &   \textbf{0.863} \\
MLP &  0.0001 &     \textbf{0.855} &    \textbf{0.359} & 0.914 &      0.861 &   0.844 \\
SPN &  0.0001 &     0.825 &    2.233 & 0.893 &      0.823 &   0.826 \\

\bottomrule
\end{tabular}

\end{table*}

\subsection{Counterfactual Evaluation}
\begin{table*}%[h!]
\caption{Counterfactual statistics of the MLP baseline classifier and the SPN for different $\beta_{1}$. For each VAE configuration separated within the rows, the influence of different $\beta$ and $\gamma$ that regularize the latent counterfactual, are shown. Best results are displayed as bold.}% (test set 1) - No fine tuning on the SPN}
\label{tab:cf_metrics}
\centering
\begin{tabular}{cccc|cccc}
\toprule
$\beta_{1}$ &Classifier&  $\beta$ &  $\gamma$ &  Validity &    L2 &    FID &  Switch epoch \\
\midrule
0.1000 &        MLP &   0.0 &    0.0 &      \textbf{1.00} & 25.54 & 398.06 &        \textbf{101.95} \\
0.1000 &        MLP &   1.0 &    0.0 &      \textbf{1.00} & 25.53 & 398.37 &        102.19 \\
0.1000 &        SPN &   0.0 &    0.0 &      \textbf{1.00} & 25.59 & 394.37 &        119.09 \\
0.1000 &        SPN &   0.0 &    1.0 &      \textbf{1.00} & 25.60 & 394.57 &        130.39 \\
0.1000 &        SPN &   1.0 &    0.0 &      \textbf{1.00} & 25.56 & 393.99 &        119.01 \\
0.1000 &        SPN &   1.0 &    1.0 &      \textbf{1.00} & 25.59 & 394.20 &        131.47 \\
\midrule
0.0100 &        MLP &   0.0 &    0.0 &      0.09 & 20.39 & 372.06 &        465.20 \\
0.0100 &        MLP &   1.0 &    0.0 &      0.09 & 20.39 & 372.07 &        470.20 \\
0.0100 &        SPN &   0.0 &    0.0 &      0.55 & 20.78 & 376.54 &        612.69 \\
0.0100 &        SPN &   0.0 &    1.0 &      0.53 & 20.78 & 375.81 &        616.27 \\
0.0100 &        SPN &   1.0 &    0.0 &      0.56 & 20.71 & 378.06 &        648.04 \\
0.0100 &        SPN &   1.0 &    1.0 &      0.51 & 20.77 & 376.55 &        601.23 \\
\midrule
0.0010 &        MLP &   0.0 &    0.0 &      0.08 & 17.29 & 324.24 &        483.25 \\
0.0010 &        MLP &   1.0 &    0.0 &      0.08 & 17.29 & 324.24 &        487.50 \\
0.0010 &        SPN &   0.0 &    0.0 &      0.89 & 17.88 & 325.80 &        451.99 \\
0.0010 &        SPN &   0.0 &    1.0 &      0.86 & 17.90 & 325.93 &        463.05 \\
0.0010 &        SPN &   1.0 &    0.0 &      0.88 & 17.86 & 326.38 &        446.24 \\
0.0010 &        SPN &   1.0 &    1.0 &      0.87 & 17.89 & 325.39 &        472.82 \\
\midrule
0.0001 &        MLP &   0.0 &    0.0 &      0.03 & \textbf{16.78} & \textbf{299.24} &        681.00 \\
0.0001 &        MLP &   1.0 &    0.0 &      0.03 & \textbf{16.78} & \textbf{299.24} &        689.67 \\
0.0001 &        SPN &   0.0 &    0.0 &      0.55 & 17.35 & 300.08 &        551.76 \\
0.0001 &        SPN &   0.0 &    1.0 &      0.44 & 17.29 & 300.90 &        544.50 \\
0.0001 &        SPN &   1.0 &    0.0 &      0.55 & 17.34 & 300.28 &        559.15 \\
0.0001 &        SPN &   1.0 &    1.0 &      0.43 & 17.27 & 301.21 &        537.53 \\

\bottomrule
\end{tabular}

\end{table*}

Table~\ref{tab:cf_metrics} presents the results of the counterfactual evaluation metrics for the MLP and SPN under varying $\beta_{1}$ and hyperparameters $ \beta $ and $ \gamma $ of Equation~\ref{eq:clf}. With $\beta_{1}=0.1$, both MLP and SPN achieve a validity of 1.00 across all settings, indicating that under strong regularization, all generated counterfactuals successfully alter model predictions. However, under weaker regularization, validity significantly declines for the MLP (0.09-0.03), while the SPN retains moderate validity (0.43-0.89). This suggests that counterfactual generation requires larger changes for the model predictions to be altered when DKL regularization is low. Additionally, the MLP's low entropy, which leads to more confident predictions, is more challenging to modify in the latent space.

Lower values of the L2 norm indicate counterfactuals that are closer to the original input. $\beta_{1}$ has the strongest influence on this metric, as it also impacts reconstruction performance, as demonstrated in Table~\ref{table:VAE_performance}. The same trend holds for FID, measuring counterfactual plausibility. Given that both L2 and FID are influenced by the overall reconstruction performance of the VAE, it will be important to consider alternative formulations of these metrics. Specifically, relative versions of L2 and FID could provide a clearer assessment of counterfactual quality by measuring either the relative change to the original reconstruction or, in case of FID, by considering the individual class distributions.

Within the individual VAE configuration, the impact of $ \beta $ and $ \gamma $ remains unexpectedly small. This observation further supports the assumption that modifying L2 and FID to better reflect counterfactual-specific characteristics might yield more informative insights. For small $\beta_{1}$, the results on the MLP baseline suggest that enabling larger changes in the latent representation $ z $ could enhance the effectiveness of the counterfactual generation objective described in Equation~\ref{eq:clf}. 

\begin{figure*}[t!]%[h!]

\begin{center}

\includegraphics[width=0.75\textwidth]{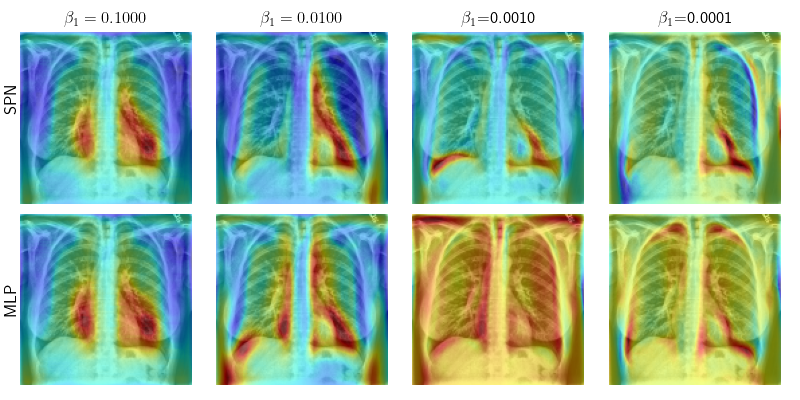}
\caption{Example visualization of input alterations in the generated counterfactuals for the different VAE regularizations with the SPN and MLP classifier. The color coding represents $\frac{1}{R}\left(\sum x_{cf_r} - \tilde{x}_{r}\right)$, where red regions indicate additions by the counterfactual and blue regions indicate reductions.}
\label{fig:vizu}
\end{center}
\end{figure*}

Figure~\ref{fig:vizu} displays an example of the mean input alteration $\frac{1}{R}\sum( x_{cf_r}-\tilde{x}_{r})$ of the generated counterfactual explanations. Red regions indicate additions and blue regions indicate reductions made by the counterfactual explanation. 
High KLD regularization with a large $\beta_1$ results in larger regions of modification. In contrast, with lower KLD regularization, the modifications become more fine-grained. When comparing the differences between the MLP and SPN classifiers, it shows that MLPs provide convincing explanations only under high KLD regularization. In contrast, SPNs consistently identify the region around the heart across all regularization settings.

Our findings highlight the trade-offs between regularization strength, counterfactual validity, proximity, and plausibility, suggesting that further refinement of evaluation metrics will be necessary to fully capture the quality of generated counterfactuals.

\section{Conclusions}
In this paper, we investigated the integration of SPNs into the latent space of VAEs, to act as classifier and regularizer for counterfactual generation. We compared our approach with a standard neural network based MLP baseline. The results highlight the impact of KLD regularization on counterfactual validity, proximity, and plausibility. Under strong regularization, counterfactuals reliably alter model predictions, whereas weaker regularization improves plausibility and proximity at the cost of reduced validity, particularly for the MLP baseline. The SPN consistently maintains at least moderate validity across all settings, demonstrating larger robustness in counterfactual generation. Visualizing the changes between original reconstructions and counterfactual reconstructions show promising results for visual decision explanations.
So far, our evaluation has focused on the mean counterfactual result for an entire sample. Future work should explore more sophisticated weighted combinations, assess the diversity of individual explanations, and analyze the most effective counterfactual explanations.
To further refine our approach, we will enhance counterfactual evaluation metrics to better capture key properties. Alternative formulations of L2 and FID, incorporating relative reconstruction changes or class-specific distributions, could provide deeper insights into counterfactual plausibility and proximity. Finally, identifying the key factors that govern latent space manipulation, particularly the trade-off between validity and plausibility, may further improve the quality and interpretability of generated counterfactuals.

\begin{acks}
The authors gratefully acknowledge the computing time granted on the supercomputer MOGON II at Johannes Gutenberg University Mainz (hpc.uni-mainz.de). The project was funded by the Federal Ministry of Research, Technology and Space (BMFTR) with the grant number 03ZU1202JB.
\end{acks}

%\printbibliography

%\bibliographystyle{ACM-Reference-Format}
%\bibliography{sample}
%\nocite{*}
%\bibliography{BioKDD2025_main}%{sample-base}
\input{BioKDD2025_main.bbl}

\end{document}